\documentclass{article}

\PassOptionsToPackage{numbers, compress}{natbib}

\usepackage[preprint]{neurips_2024}

\usepackage[utf8]{inputenc} 
\usepackage[T1]{fontenc}    
\usepackage{hyperref}       
\usepackage{url}            
\usepackage{booktabs}       
\usepackage{amsfonts}       
\usepackage{nicefrac}       
\usepackage{microtype}      
\usepackage{xcolor}         
\usepackage{amsmath}
\usepackage{graphicx}
\usepackage{lipsum}
\usepackage{caption}
\usepackage{multirow}
\usepackage{makecell}
\usepackage{adjustbox}
\usepackage{wrapfig}
\usepackage{cleveref}
\usepackage{pifont}
\usepackage{xfrac}
\usepackage{float}  
\newcommand{\expec}{\mathbb{E}}
\newcommand{\ie}{\emph{i.e.}}    
\newcommand{\eg}{\emph{e.g.}}    
\newcommand{\encoder}{\mathcal{E}}
\newcommand{\model}{\epsilon_\theta}
\newcommand{\conditioner}{\tau_\theta}

\newlength\savewidth

\title{
ResMaster: Mastering High-Resolution Image Generation via Structural and Fine-Grained Guidance}

\author{%
  Shuwei Shi$^{1}$, Wenbo Li$^{2}$, Yuechen Zhang$^{2}$, Jingwen He$^{2}$, Biao Gong$^{3}$, Yinqiang Zheng$^{1}$\thanks{Corresponding author.}\\
  $^{1}$~The University of Tokyo
  $^{2}$~The Chinese University of Hong Kong
  $^{3}$~Ant Group \\
  \texttt{\{shishuwei666, fenglinglwb\}@gmail.com, yqzheng@ai.u-tokyo.ac.jp} 
}

\begin{document}

\maketitle

\begin{figure}[h]
    \centering
    \includegraphics[width=1.\linewidth]{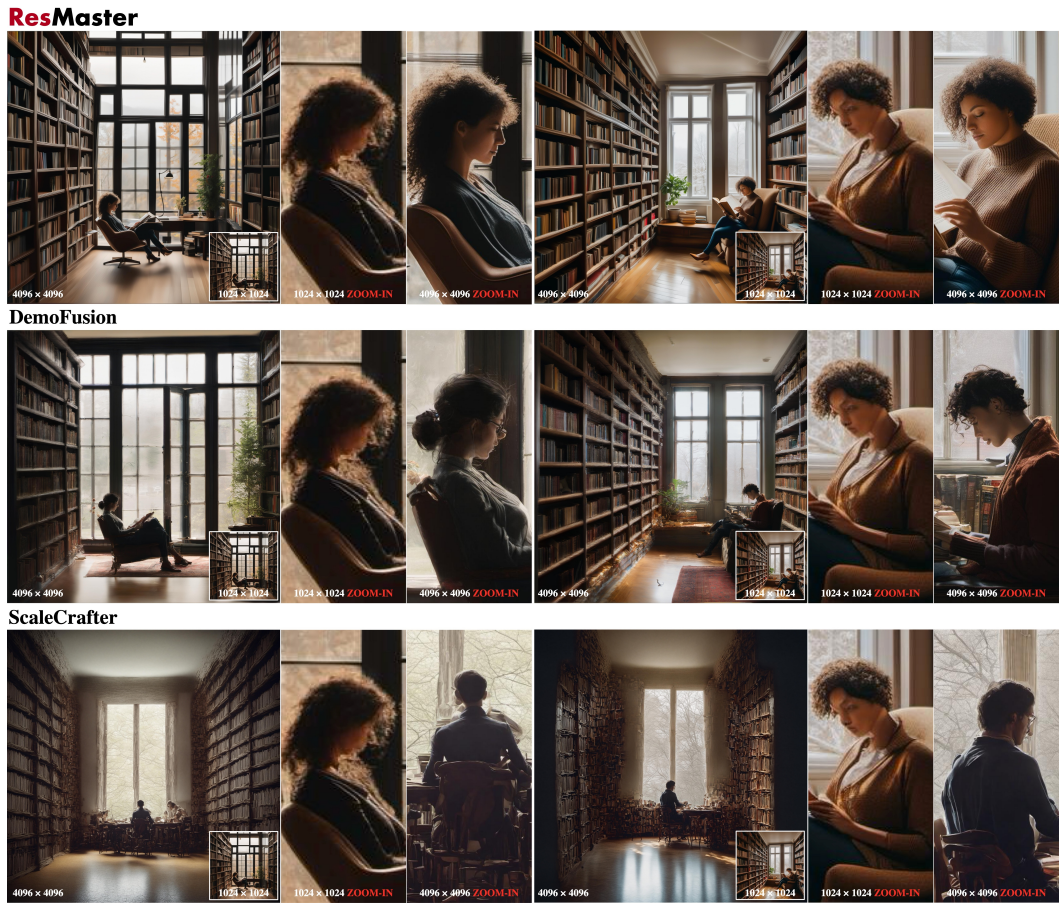}
    \vspace{-3mm}
    \caption{\textbf{Comparisons of $16\times$ ($4096 \times 4096$) image generation based on SDXL~\cite{podell2023sdxl}.} Our ResMaster can restore lost details and complex structures (\textit{e.g.}, faces and hands) from low-resolution originals while preserving structural integrity and semantic fidelity compared to ScaleCrafter~\cite{he2023scalecrafter} and DemoFusion~\cite{du2024demofusion}. We upscale the $1024\times1024$ image to the same resolution to facilitate comparison.}
    \label{fig:teaser}
\end{figure}

\begin{abstract}
Diffusion models excel at producing high-quality images; however, scaling to higher resolutions, such as 4K, often results in over-smoothed content, structural distortions, and repetitive patterns. To this end, we introduce ResMaster, a novel, training-free method that empowers resolution-limited diffusion models to generate high-quality images beyond resolution restrictions. 
Specifically, ResMaster leverages a low-resolution reference image created by a pre-trained diffusion model to provide structural and fine-grained guidance for crafting high-resolution images on a patch-by-patch basis. 
To ensure a coherent global structure,
ResMaster meticulously aligns the low-frequency components of high-resolution patches with the low-resolution reference at each denoising step. For fine-grained guidance, tailored image prompts based on the low-resolution reference and enriched textual prompts produced by a vision-language model are incorporated. This approach could significantly mitigate local pattern distortions and improve detail refinement. Extensive experiments validate that ResMaster sets a new benchmark for high-resolution image generation and demonstrates promising efficiency. The project page is \url{https://shuweis.github.io/ResMaster}.
\end{abstract}

\begin{figure}[pb!]
    \centering
    \includegraphics[width=1.\linewidth]{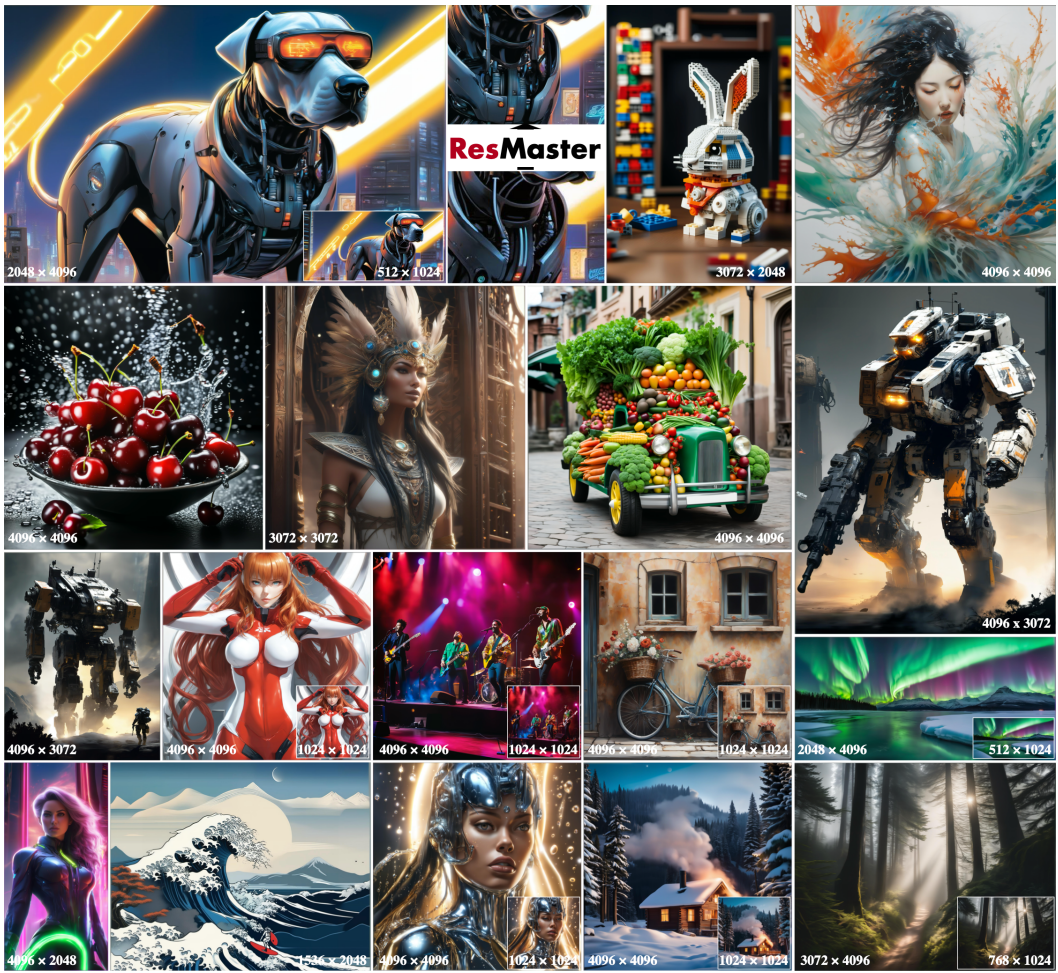}
    \vspace{-3mm}
    \caption{\textbf{Multi-aspect-ratio images generated by ResMaster versus SDXL~\cite{podell2023sdxl}.} SDXL can synthesize high-quality $1024 \times 1024$ images. ResMaster can further upscale the generated results by 16 times or more without retraining the text-to-image diffusion model. \textbf{Best viewed ZOOMED-IN.}}
    \label{fig:teaser2}
\end{figure}

\section{Introduction}
In recent years, diffusion models~\cite{chefer2023attend,epstein2023diffusion,rombach2022high,chen2024training,peebles2023scalable,brooks2023instructpix2pix, mou2023dragondiffusion} have significantly advanced the fields of image generation, drawing considerable attention from the research community. Although representative models such as Stable Diffusion XL~\cite{podell2023sdxl} (SDXL), DALL·E 3~\cite{Dalle-3} and Midjourney~\cite{Midjourney} can generate high-quality images, they perform well only within resolutions of $1024\times1024$. This limitation hinders their applications that require generated images with higher resolutions (\eg, 4k).

Several methods \cite{bar2023multidiffusion,he2023scalecrafter,du2024demofusion} achieve higher-resolution image generation by adapting pre-trained diffusion models (\eg  SDXL) without additional re-training. The pioneering work, Multi-Diffusion~\cite{bar2023multidiffusion}, generates higher-resolution images by stitching generated high-resolution overlapping patches. 
However, this approach cannot guarantee global consistency, since there is no explicit structural guidance during generation.
Moreover, it often results in repetitive objects that correspond to the provided prompt due to patch-wise generation.
To alleviate the above issues, 
ScaleCrafter~\cite{he2023scalecrafter} adopts whole image generation instead of the patch-wise counterpart. Specifically, it replaces some convolutions with dilated convolutions to preserve global consistency. However, this modification in convolution layers hampers the generative ability inherited from pre-trained diffusion models, leading to obvious structural distortions and repetitive local patterns, as demonstrated in \figurename~\ref{fig:teaser}. On the other hand, DemoFusion~\cite{du2024demofusion} introduces dilated sampling to ensure global consistency. Nonetheless, due to the lack of fine-grained guidance, repetitive patterns continue to plague this approach, as shown in \figurename~\ref{fig:teaser}. In summary, existing methods struggle to achieve a good balance between global consistency and reasonable local detail generation.

To this end, we propose \textbf{ResMaster}, a novel higher-resolution image generation method that employs \textit{Structural and Fine-Grained Guidance} to ensure structural integrity and enhance detail generation.
Specifically, ResMaster implements low-frequency component swapping using the low-resolution image generated at each sampling step to maintain global structural coherence in higher-resolution outputs. Additionally, to mitigate repetitive patterns and increase detail accuracy, we employ localized fine-grained guidance using condensed image prompts and enriched textual descriptions. The image prompts, derived from the generated low-resolution counterparts, contain critical semantic and structural information. Simultaneously, the detailed textual prompts produced by a pre-trained vision-language model (VLM) contribute to image generation on more complex and accurate patterns.

With these techniques, ResMaster is able to generate high-quality higher-resolution images across various aspect ratios (see \figurename~\ref{fig:teaser} and \figurename~\ref{fig:teaser2}). The generated images exhibit both reasonable structure and photo-realistic details. Extensive qualitative and quantitative experiments demonstrate that ResMaster delivers superior results compared to other higher-resolution image generation methods.

\section{Related Work}
\label{related_work}
\paragraph{Text-to-Image Diffusion Models.} 
Text-to-image diffusion models~\cite{dhariwal2021diffusion, feng2023ranni, ho2022cascaded, ho2020denoising, Song2020DenoisingDI, nichol2021improved, ramesh2022, stablediffusion2} represent a cutting-edge advancement in generative technology, leveraging the power of diffusion probabilistic models~\cite{ho2020denoising, song2020denoising} to synthesize high-quality images from textual descriptions. These models operate by gradually denoising a noisy input through iterative refinement steps. This process is guided by conditioning information that typically comes from textual input, allowing the generation of detailed and contextually relevant imagery. Furthermore, Latent Diffusion Models (LDMs)~\cite{rombach2022high} perform a similar process within a compact latent space, improving both the efficiency and scalability of the model and maintaining high fidelity in the generated images. This approach has been further refined in SDXL~\cite{podell2023sdxl} and other models~\cite{zheng2024cogview3, teng2023relay, huang2023learning, lu2024fit, zhu2024sd}, which are applied in creative fields. However, despite these advancements, challenges remain, particularly in achieving higher resolution output (e.g., 4K) without compromising the generative quality.
\paragraph{High-Resolution Image Synthesis.} 
High-resolution image generation has attracted significant attention from researchers. The primary challenges of this task are learning the correct distributions from high-dimensional data and the immense consumption of computational resources. Previous studies can be categorized into training-based methods~\cite{zheng2024any, teng2023relay, chen2023pixart} and training-free methods~\cite{bar2023multidiffusion, he2023scalecrafter, lee2023syncdiffusion, gong2024check}. Training-based methods are trained on specific sets of images within the range of target resolution and aspect ratio. For example, Any-size-Diffusion~\cite{zheng2024any} leverages a selected set of images with a restricted range of ratios to optimize the diffusion model. PixArt-$\Sigma$~\cite{chen2024pixart} uses a weak-to-strong training strategy to train their model on higher-quality datasets. However, these methods are still limited by the resolution of their training images and cannot generate high-quality images beyond resolution restrictions. In contrast, training-free strategies aim to utilize pre-trained diffusion models without additional training phases. For instance, MultiDiffusion~\cite{bar2023multidiffusion} addresses high-resolution synthesis by fusing multiple denoising paths. SyncDiffusion~\cite{lee2023syncdiffusion} further improves the consistency among different patches through gradient-guided manipulation in the process of denoising. However, these methods suffer from local repetitions and distortion of object structures. To alleviate the aforementioned issues, ScaleCrafter \cite{he2023scalecrafter} employs dilated convolutions to increase the receptive field of the convolution kernel. This has been proven to be effective in eliminating the repetitiveness of the object's structure. However, when applied to higher-resolution image generation tasks, issues such as structural distortion and a decline in the quality of local detail generation begin to emerge. DemoFusion~\cite{du2024demofusion} ensures the accuracy of target structures and produces more details through skip residual and dilated sampling. It shows further improvement in generating high-quality images, but it is still affected by repetitive objects and chaotic local details. In this paper, we propose a method with structural and fine-grained guidance, which ensures structural accuracy while enhancing the details of high-resolution images, alleviating the issues mentioned above.

\section{Methodology}
\label{method}
\subsection{Preliminaries}
\paragraph{Latent Diffusion Model.} Our methods are built on the forefront text-to-image diffusion model, SDXL~\cite{podell2023sdxl}, which belongs to the series of LDMs. Given an image $x\in\mathbb{R}^{H \times W \times 3}$, the encoder $\encoder$ in LDM first encodes it into latent representation $z=\encoder(x)$, where $z\in\mathbb{R}^{\sfrac{H}{8} \times \sfrac{W}{8} \times C}$. Then, forward diffusion and reverse denoising are conducted in the latent space. In the forward process, the noise is gradually added to the latent $z$ within $T$ steps, represented as
\begin{equation}
q\left(\boldsymbol{z}_{t} \mid \boldsymbol{z}_{t-1}\right)=N\left(\boldsymbol{z}_{t} ; \sqrt{1-\beta_{t}} \boldsymbol{z}_{t-1}, \beta_{t} \boldsymbol{I}\right),
\label{eq:diffprocess}
\end{equation}
where $\beta_{t}$ is the variance schedule, and $t\in \{1,...,T\}$. On the other hand, in the backward process, a Unet $\epsilon_\theta$ is used to predict the noise iteratively, eventually yielding results under the guidance of the text prompt $y$. The object of this stage can be formulated as:
\begin{equation}
{L} = \expec_{\encoder(x), y, \epsilon \sim \mathcal{N}(0, 1), t }\Big[ \Vert \epsilon - \model(z_{t},t, \conditioner(y)) \Vert_{2}^{2}\Big] \, ,
\label{eq:revprocess}
\end{equation}
where $\conditioner$ is the text encoder of CLIP~\cite{radford2021learning}.
\paragraph{Patch-based Diffusion Model.}
Multi-Diffusion~\cite{bar2023multidiffusion} initially generates higher-resolution images through a denoising process that utilizes overlapping patches. This approach is widely adopted in subsequent works~\cite{lee2023syncdiffusion, du2024demofusion} due to its flexibility and convenience.  Specifically, given a latent representation $z_{t}\in\mathbb{R}^{\sfrac{H}{8} \times \sfrac{W}{8} \times C}$, it is first partitioned into patches $z_{n,t}\in\mathbb{R}^{h \times w \times C}$ with a specified window size [$h, w$] and stride [$d_{h}, d_{w}$], resulting in a total of $N=\left(\frac{(\sfrac{H}{8}-h)}{d_{h}}+1\right) \times\left(\frac{(\sfrac{W}{8}-w)}{d_{w}}+1\right)$ patches. Each patch is individually denoised, with overlapping areas averaged at each step.
\subsection{Model Framework}
\begin{figure}[!t]
    \centering
    \includegraphics[width=1.\linewidth]{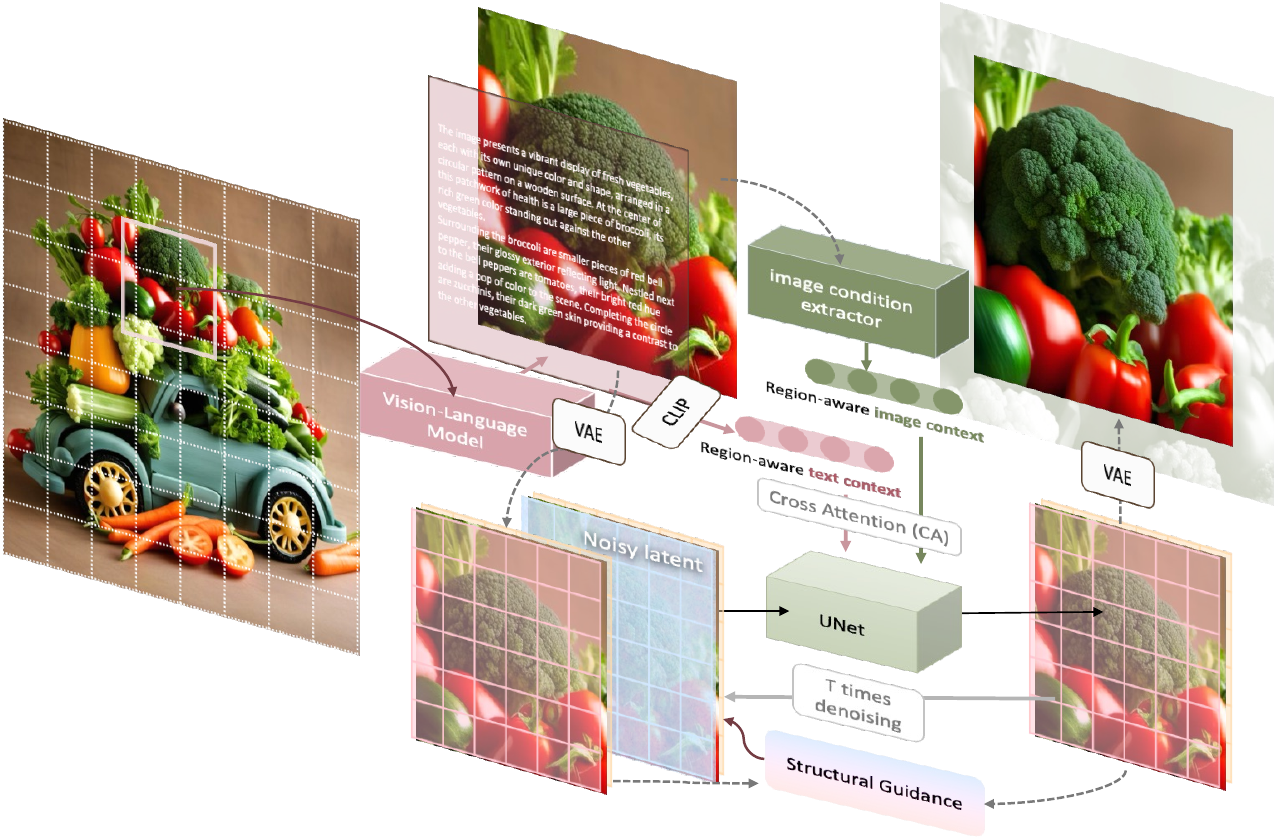}
    \caption{\textbf{The overall framework of ResMaster.} ResMaster is a patch-based denoising diffusion model that includes structural and fine-grained guidance. Fine-grained guidance utilizes an Image Condition Extractor and a Vision-Language Model to extract region-aware image features and re-caption text prompts, respectively. These conditions are then used together via Cross Attention to guide the denoising process of the current patch. Furthermore, structural guidance ensures the structure of the generated image through low-frequency component swapping.}
    \label{fig:pipeline}
\end{figure}

Our ResMaster generates high-resolution images guided by their low-resolution counterparts. As shown in \figurename~\ref{fig:pipeline}, we first use SDXL to create a low-resolution image $\mathbf{I}^L$ based on the prompt $p$. This image is then upsampled to the target resolution using bicubic interpolation, resulting in $\mathbf{B}^L$, which is divided into $N$ equal-sized overlapping patches. We then apply structural guidance and fine-grained guidance. a) \textbf{Structural Guidance}: To improve structural coherence, we use the VAE encoder to transform the $i$-th low-resolution patch of $\mathbf{B}^L$ into $\mathbf{z}_i^{L}$, which serves as a guidance map. The $i$-th high-resolution noise patch at time $t$, dubbed $\mathbf{z}_{i,t}^H$, is mapped to a preliminary estimate $\mathbf{z}_{i,0 \mid t}^H$. We align the low-frequency components of $\mathbf{z}_{i,0 \mid t}^H$ with $\mathbf{z}_i^{L}$ for structural guidance. b) \textbf{Fine-grained Guidance}: Each low-resolution patch of $\mathbf{B}^L$ is processed by the Image Condition Extractor and the Large Vision Language Model to yield fine-grained image and textual representations. These representations are injected into the generative network via cross-attention to guide the noise prediction more accurately.

\subsection{Structural Guidance}
\begin{figure}[!t]
    \centering
    \includegraphics[width=1.\linewidth]{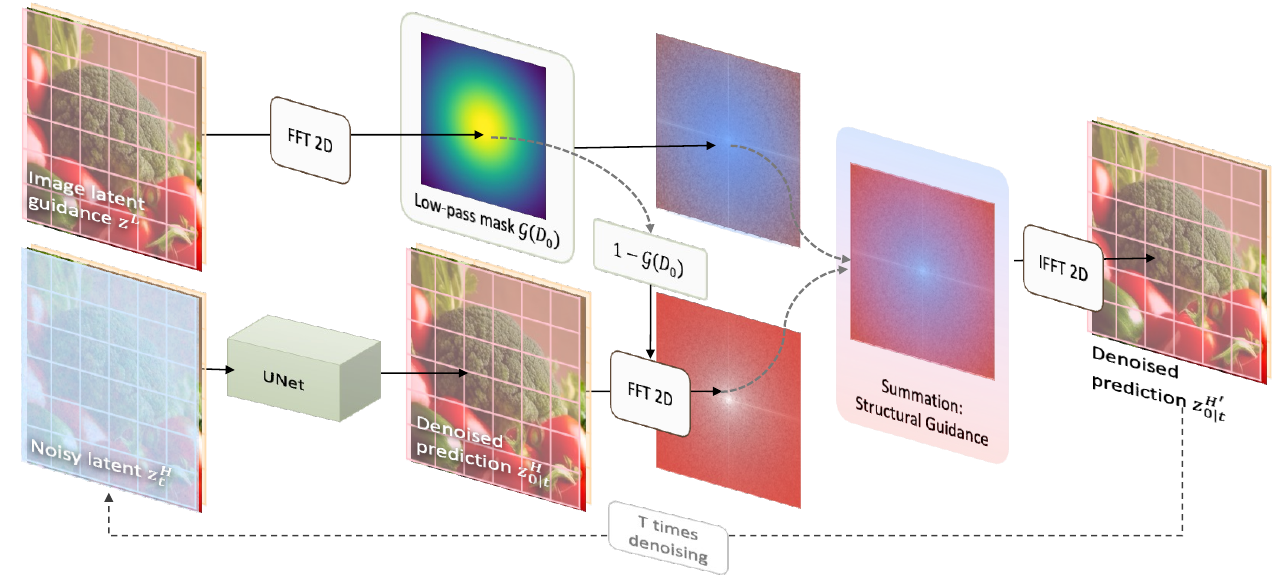}
    \caption{\textbf{The overall pipeline of Structural Guidance.} We use 2D Fast Fourier Transform to convert images to the frequency domain and apply a Gaussian low-pass filter to extract low-frequency information for exchange. This low-frequency information is then fused with the original high-frequency information and converted back to the spatial domain.}
    \label{fig:structure}
\end{figure}

Due to the distribution disparity between the training data and target high-resolution images, previous patch-based diffusion models often exhibit structural distortions and semantic confusion, impairing visual quality. To enhance structural rationality, we propose using generated low-resolution images for structural guidance. A common approach, as noted in~\cite{dhariwal2021diffusion}, updates $\mathbf{z}_{0\mid t}^H$ to align with $\mathbf{z}^{L}$ via gradient decay. However, this method increases time and memory consumption and introduces blurriness from the upsampled low-resolution image into the generated result. To mitigate these issues, we propose low-frequency (\ie, structure~\cite{ren2024consisti2v}) component swapping, as illustrated in Figure~\ref{fig:structure}, which is efficient and resource-friendly. At time step $t$, we predict $\mathbf{z}_{0 \mid t}^H$ from $\mathbf{z}_{t}^H$ and replace its low-frequency components with those of $\mathbf{z}^{L}$, resulting in $\mathbf{z}_{0 \mid t}^{H^\prime}$. This ensures proper structural guidance, formulated as follows:
\begin{align}
    \mathcal{F}^{low}_{\mathbf{z}^{L}} &= \texttt{FFT\_2D}(\mathbf{z}^{L})\odot \mathcal{G}(D_0) \,, \\
    \mathcal{F}^{high}_{\mathbf{z}_{0\mid t}^H} &= \texttt{FFT\_2D}(\mathbf{z}_{0\mid t}^H)\odot \left(1 - \mathcal{G}(D_0) \right) \,, \\
    \mathbf{z}_{0\mid t}^{H^\prime}&= \texttt{IFFT\_2D}(\mathcal{F}^{low}_{\mathbf{z}^{L}} + \mathcal{F}^{high}_{\mathbf{z}_{0\mid t}^H}) \,,
\end{align}
where \texttt{FFT\_2D} is the 2D Fast Fourier Transform, and \texttt{IFFT\_2D} is its inverse. $\mathcal{G}$ represents the Gaussian low-pass filter, and $D_0$ is the normalized cutoff frequency. The term $\mathbf{z}_{0\mid t}^H$ is calculated by:
\begin{equation}
\mathbf{z}_{0\mid t}^H \approx \left(\mathbf{z}_{t}^H-\sqrt{1-\bar{\alpha}_{t}} \boldsymbol{\epsilon}_{\theta}\left(\mathbf{z}_{t}^H \right)\right) / \sqrt{\bar{\alpha}_{t}} \,, \\
\end{equation}
where $\boldsymbol{\epsilon}_{\theta}$ is the denoising Unet, ${\alpha}_{t}$ is the prescribed variance schedule and $\bar{\alpha}_{t} = \prod_{i=1}^{t} \alpha_{i}$. Then, the final result $z_{t-1}^H$ is derived as:
\begin{equation}
q\left(\mathbf{z}_{t-1}^H \mid \mathbf{z}_{t}^H, \mathbf{z}_{0\mid t}^{H^\prime} \right)=\mathcal{N}\left(\mathbf{z}_{t-1}^H ; \tilde{\boldsymbol{\mu}}_{t}\left(\mathbf{z}_{t}^H, \mathbf{z}_{0\mid t}^{H^\prime} \right), \tilde{\beta}_{t} \mathbf{I}\right) \,.
\end{equation}
This method maintains reasonable global structures in high-resolution images while retaining the generative ability of diffusion models. However, structural guidance alone may not suffice to generate locally accurate and rich details, necessitating fine-grained guidance.
\subsection{Fine-Grained Guidance}
\paragraph{Tailored image prompts.} To further enhance the consistency of local structure and mitigate the repeated patterns resulting from the identical text prompt across different patches~\cite{bar2023multidiffusion,du2024demofusion}, we customize multiple image prompts, inspired by previous successful methodologies~\cite{ye2023ip-adapter, sun2023coser}. These methods have been proven to generate images consistent with specific attributes based on the information from the input images. First, a CLIP image encoder extracts the class token from the low-resolution image, mapping it to representative image features. Decoupled cross-attention mechanisms are then employed to integrate these image and text features into the pre-trained text-to-image diffusion model, formulated as:
\begin{equation}
\mathbf{X}^{\prime} = \operatorname{Softmax}\left(\frac{\mathbf{Q} \mathbf{K}_{\rm{T}}^{\top}}{\sqrt{d}}\right) \mathbf{V}_{\rm{T}} + \lambda \operatorname{Softmax}\left(\frac{\mathbf{Q} \mathbf{K}_{\rm{I}}^{\top}}{\sqrt{d}}\right) \mathbf{V}_{\rm{I}} \,,
\end{equation}
where $\mathbf{Q}=\mathbf{X} \mathbf{W}_{q}$, $\mathbf{K}_{\rm{T}}=\boldsymbol{c}_{\rm{T}} \mathbf{W}_{k}$, $\mathbf{V}_{\rm{T}}=\boldsymbol{c}_{\rm{T}} \mathbf{W}_{v}$, $\mathbf{K}_{\rm{I}}=\boldsymbol{c}_{\rm{I}} \mathbf{W}_{k}^{\prime}$, and $\mathbf{V}_{\rm{I}}=\boldsymbol{c}_{\rm{I}} \mathbf{W}_{v}^{\prime}$ represent the query, key, and values of text and image features respectively, and $\lambda$ is the weighting factor.

This design, with patch-wise image prompts, significantly alleviates the issue of repeated patterns, resulting in improved local structures. Nevertheless, due to the lack of details in low-resolution images, the extracted image prompts only provide structural guidance and cannot fully activate the network's generative capabilities. Thus, more informative text prompts are needed. Existing methods using global text information face two issues. First, the patch-based method does not align well with global text. Second, global text often lacks detailed descriptions of local features. Therefore, incorporating more detailed descriptions is essential, as evidenced by existing text-to-image methods~\cite{chen2024pixart,betker2023improving}.
\paragraph{Enriched textual prompts.}
To obtain more detailed and accurate descriptions for image patches, we introduce a pre-trained large Vision-Language Model (VLM), Share-Captioner~\cite{chen2023sharegpt4v}, to re-caption low-resolution image patches. This model has proven to be effective in better aligning textual and visual information and reducing hallucinations~\cite{chen2024pixart}. Our experiments indicate that providing the full image's description as context to the VLM does not yield additional benefits and may lead to hallucinations. Therefore, the instruction we give to the VLM is ``\textit{Describe the following image patch in detail.}'' Following this instruction, the VLM generates a detailed prompt for each patch. These enriched prompts enable our method to produce more accurate structures and richer local details, resulting in higher visual quality.
\section{Experiments}
\label{experiments}
In this section, we report the qualitative and quantitative results and ablation studies. We validate the performance of ResMaster based on the SDXL~\cite{podell2023sdxl}. More qualitative results can be found in \figurename~\ref{fig:appendix_qual} and \figurename~\ref{fig:appendix1}.
\subsection{Comparison}
We compare our method with the following representative generative approaches:
(i) SDXL~\cite{podell2023sdxl} Direct Inference, which uses pre-trained SDXL to directly infer the target resolution images. Here, the resolution of the results we compare is higher than its training resolution.
(ii) SDXL+BSRGAN. We use the classic super-resolution method BSRGAN~\cite{zhang2021designing} to conduct image super-resolution. This is a traditional approach to increase image resolution. (iii) SCALECRAFTER~\cite{he2023scalecrafter}, a method that generates high-resolution images directly using dilated convolutions or large convolutional kernels. (iv) DemoFusion~\cite{du2024demofusion}, a high-resolution image generation method based on MultiDiffusion~\cite{bar2023multidiffusion}, using dilated sampling to ensure global structural consistency.

\begin{figure}[!t]
    \centering
    \includegraphics[width=1.\linewidth]{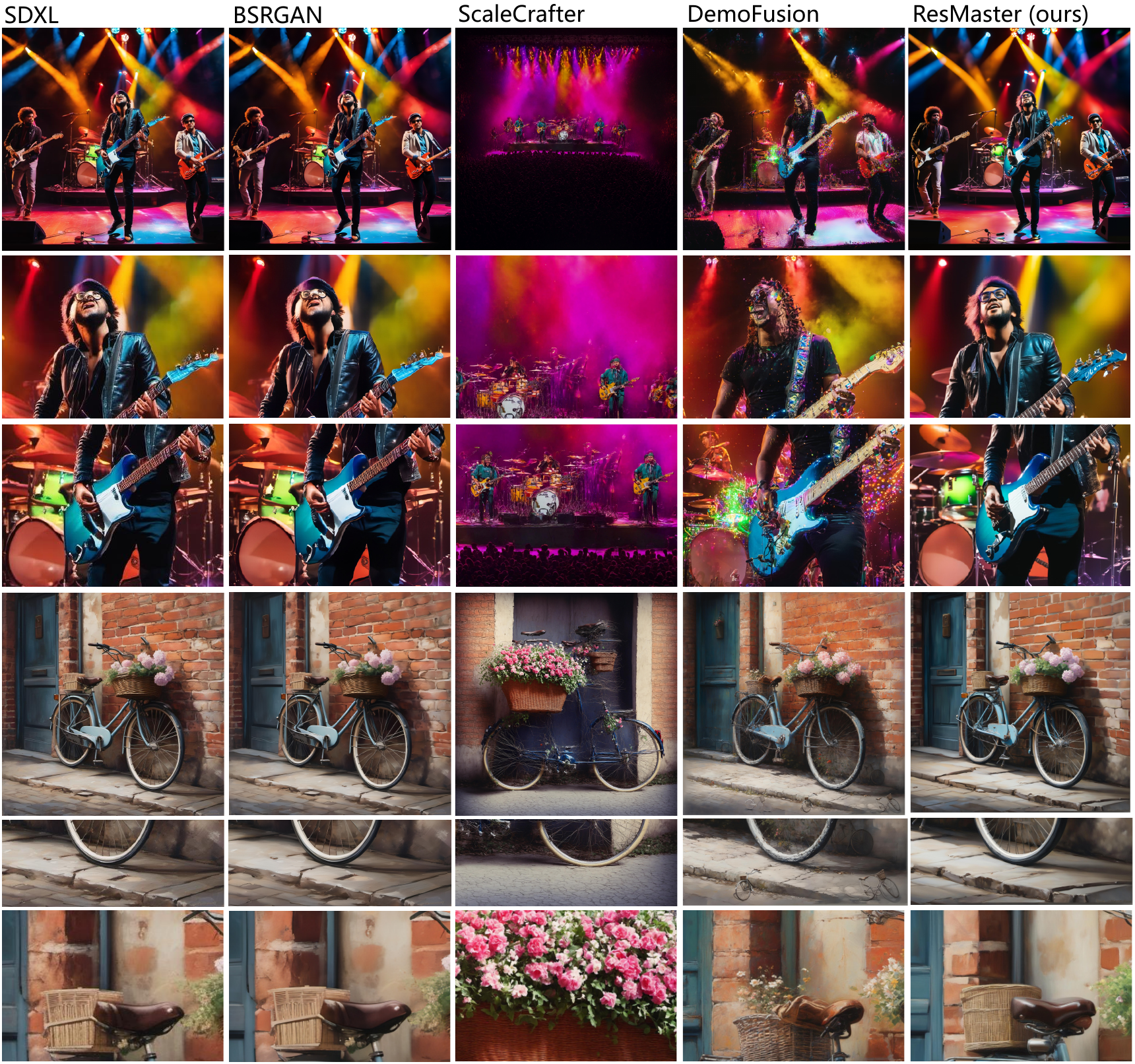}
    \caption{\textbf{Qualitative comparisons with other methods.} All results are presented at a resolution of $4096 \times 4096$ ($16 \times$), with the SDXL results being directly upscaled from $1024 \times 1024$. Some areas have been zoomed in.}
    \label{fig:compare}
\end{figure}

\begin{figure}[!t]
    \centering
    \includegraphics[width=1.\linewidth]{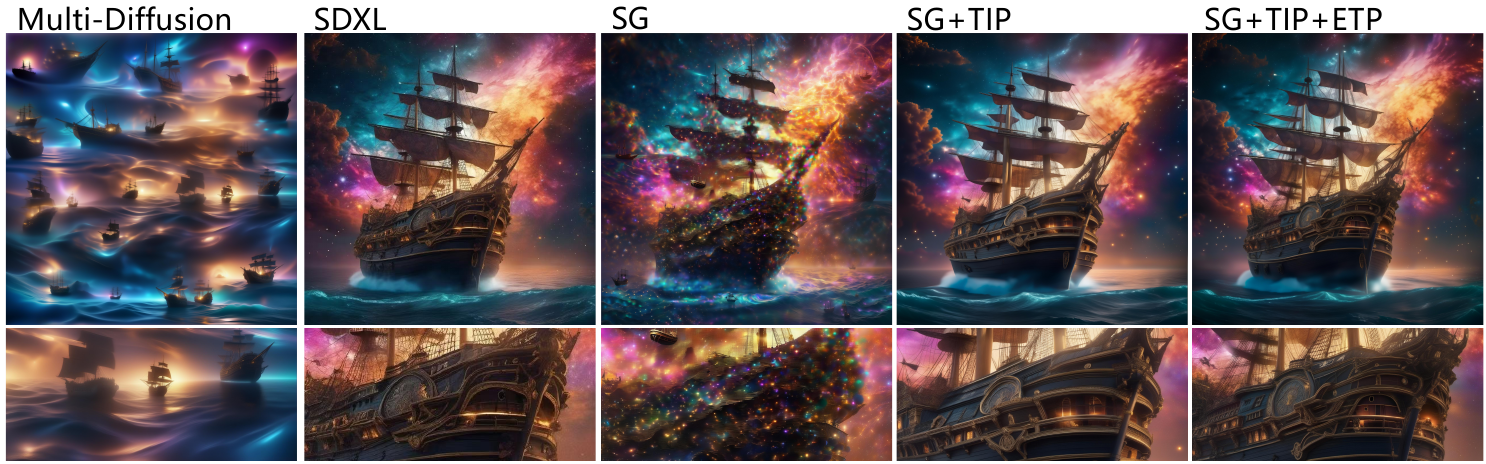}
    \caption{\textbf{The Ablation study of three components used in ResMaster}: Structural Guidance (SG), Tailored Image Prompts (TIP) and Enriched Textual Prompts (ETP). All results are presented at a resolution of
4096 × 4096 (16×), with the SDXL results being directly upscaled from 1024 × 1024.}
    \label{fig:ablation}
\end{figure}

\subsection{Qualitative Results}
\figurename~\ref{fig:compare} shows a visual comparison of different models, each producing $16\times$ resolution results ($4096 \times 4096$) compared to the original SDXL. We select complex real-world scenes and the example prone to mode confusion to demonstrate the superiority of our method. Firstly, in the case of the first complex scene, the image generated by SDXL has a lower resolution and lacks fine-grained generation guidance for local content. As a result, the faces of people and the complex background structures are not clear and complete. BSRGAN performs super-resolution based on SDXL, partially eliminating the blurriness in upscaled results of SDXL. However, it is evident that BSRGAN merely sharpens the low-resolution results, making it unsuitable for high-resolution image generation tasks. High-resolution image generation requires more local details and, in some scenarios, the ability to correct and complete the inherent issues in low-resolution images. ScaleCrafter and DemoFusion introduced inappropriate details, resulting in incorrect textures on faces and objects. This is mainly due to ScaleCrafter using dilated convolutions, which weakens its ability to maintain structure and handle high-frequency details in high-resolution image generation. DemoFusion lacks fine-grained guidance for local content generation, leading to chaotic structures and details in the objects. In contrast, ResMaster effectively restores clear facial features and improves the structure of complex objects, making them clearer, more complete, and aesthetically pleasing. Similarly, the results in the bicycle case confirm our observations. BSRGAN still fails to add more details. ScaleCrafter has weak structure preservation and chaotic details. DemoFusion produced repeated objects, altered the seat structure, and exhibited mode confusion between different objects (flowers in the rear basket). In conclusion, our proposed ResMaster can enhance and improve more details while ensuring structural accuracy, owing to our proposed structural and fine-grained guidance. This capability is particularly crucial in high-resolution image generation.

\subsection{Quantitative Results}

\begin{table}[t]
\small
\centering
\caption{\textbf{Quantitative comparison results}. The best results are marked in \textbf{bold}, and the second best results are marked by \underline{underline}.}
\label{tab:comparison}
\scalebox{1.0}{\begin{tabular}{cl|cccccc}
\toprule
Resolution                 & Method           & FID$_r$ $\downarrow$ & IS$_r$ $\uparrow$ & FID$_c$ $\downarrow$ & IS$_c$ $\uparrow$ & CLIP $\uparrow$ & Time \\
\midrule
\multirow{5}{*}{$2048\times2048$} &  SDXL Direct Inference~\cite{podell2023sdxl} & 90.33  &  13.13   &  63.47    &  21.74   &  29.18    &  1min    \\
                           & SDXL + BSRGAN~\cite{zhang2021designing}   &   \textbf{67.00}  &  17.10   &  \underline{42.79}    &  22.36   &  \underline{31.64}    &   1min   \\
                           & SCALECRAFTER~\cite{he2023scalecrafter}     &   78.95   & 16.23    &  58.86    &  21.71   &   30.23   &   1min   \\
                           & DemoFusion~\cite{du2024demofusion}       &  69.81    & \underline{17.95}    &  45.42    &  \underline{24.53}   &   31.45   &   2min   \\
                           & ResMaster (Ours) &  \underline{68.16}    &  \textbf{18.01}   &    \textbf{40.22}  &  \textbf{24.62}   &  \textbf{31.79}    &   1min   \\ \midrule
\multirow{5}{*}{$2048\times4096$} & SDXL Direct Inference~\cite{podell2023sdxl} &  131.69    & 8.49    &   78.49   & 17.31    &   26.64   &   2min  \\
                           & SDXL + BSRGAN~\cite{zhang2021designing}    &   \textbf{69.04}   &  13.89   &  \textbf{43.50}    & 16.91    &   \underline{29.99}   &   1min   \\
                           & SCALECRAFTER~\cite{he2023scalecrafter}     &  119.67    &  8.75   &  106.15    & 11.13    &   27.53   &    3min  \\
                           & DemoFusion~\cite{du2024demofusion}       &   \underline{69.30}   & \underline{14.65}    &  49.45    &  \underline{20.16}   &   29.29   &   5min   \\
                           & ResMaster (Ours) &  69.57    & \textbf{14.67}    &   \underline{45.34}   &  \textbf{21.05}   &    \textbf{30.05}  &  3min    \\ \midrule
\multirow{5}{*}{$4096\times4096$} & SDXL Direct Inference~\cite{podell2023sdxl} &  173.42    &  7.50   &   89.46 &  16.42   &  24.54    &   5min   \\
                           & SDXL + BSRGAN~\cite{zhang2021designing}    &  \underline{67.08}    &17.14&  \textbf{50.89}  &  15.30   &  30.61    &  1min    \\
                           & SCALECRAFTER~\cite{he2023scalecrafter}     &  107.46    &  11.06   & 107.88 &  10.94   &   29.77   &   9min  \\
                           & DemoFusion~\cite{du2024demofusion}       &  76.03    &  \underline{17.85}   &  56.75   &  \underline{16.48}   &  \underline{30.83}   &  11min    \\
                           & ResMaster (Ours) &  \textbf{65.43}    &  \textbf{18.44}   &  \underline{55.09}    &  \textbf{16.51} &  \textbf{30.95}    &   6min   \\ \bottomrule
\end{tabular}}
\end{table}
For the fair evaluation of the performance of the model, we perform quantitative experiments on the dataset of Laion-5B~\cite{schuhmann2022laion} with a large number of image-caption pairs. We randomly sample $1K$ captions as the text prompts for the high-resolution image generation. Additionally, we randomly sample $10K$ images from Laion-5B as a real image set. We adopt 3 metrics following prior works~\cite{he2023scalecrafter,du2024demofusion}: Frechet Inception Distance(FID)~\cite{heusel2017gans}, Inception Score(IS)~\cite{salimans2016improved} and CLIP Score~\cite{radford2021learning} to evaluate both image quality and semantic similarity between image features and text prompts. Among them, FID$_r$ and IS$_r$ require resizing the test images to $299^2$, which may influence the evaluation results for high-resolution images. For more reasonable evaluation, we follow~\cite{zheng2024any} to crop and resize some local patches at $1K$ resolution to compute FID$_c$ and IS$_c$. We report quantitative results at three different resolutions. The inference time is performed on a single NVIDIA A100 GPU. As shown in \tablename~\ref{tab:comparison}, ResMaster has achieved state-of-the-art performance in multiple metrics. We surpass the previous state-of-the-art method, DemoFusion, in almost all metrics. DemoFusion is known for producing rich details. Due to our fine-grained guidance strategy, ResMaster achieves better detail accuracy and semantic alignment, as reflected in the improved CLIP score. In comparisons of FID metrics at various resolutions, ResMaster consistently ranks in the top two. SDXL+BSRGAN, which strictly adheres to low-resolution inputs, has been shown to suffer from blurriness in high-resolution image generation and lacks the ability to produce rich details~\cite{du2024demofusion}. Meanwhile, ResMaster ensures higher generation quality while achieving faster inference speeds compared to other representative methods. Generating a $4K$ image is 5 minutes faster than DemoFusion.
\subsection{Ablation Study}
ResMaster primarily consists of two components: structural guidance and localized fine-grained guidance. The localized fine-grained guidance includes Tailored Image Prompts and Enriched Textual Prompts. To visually present the contribution of each module, we progressively display the effects brought by the introduction of each module, as shown in \figurename~\ref{fig:ablation}. The images we present are generated at a resolution of $4096\times4096$. ResMaster is built on patch-based multi-diffusion. Therefore, when all modules are removed, the model degrades to Multi-Diffusion~\cite{bar2023multidiffusion}. We present the results of the base model, Multi-Diffusion. Without the intervention of guidance strategies, Multi-Diffusion exhibits structural distortions and repeated patterns. With the introduction of the proposed guidance strategies, structural distortions and repeated patterns disappear. Specifically, when we introduce the structural guidance strategy, the issue of structural distortions is resolved. However, due to the low resolution of the guidance images, local pattern details exhibit misalignment. We further introduce Tailored Image Prompts which can significantly alleviate the issue, making the structures and details clearer. Nonetheless, Tailored Image Prompts do not introduce additional information to supplement local details. Further incorporating Enriched Textual Prompts enhances the richness of details in high-resolution images. In conclusion, when combined, they leverage their respective strengths and functionalities, resulting in impressive generative outcomes. More quantitative results of ablation study can be found in \tablename \ref{tab:quantitative_ablation}.

\section{Limitations and Future Work} 
\label{limit}
As a training-free method, the ResMaster model is built on a pre-trained text-to-image model. Therefore, its performance is influenced by the performance of the pre-trained model. Although it has some capability for local content correction and detail generation, the final outcome is still limited by the generation results of the low-resolution image. Additionally, when generating higher-resolution images, ResMaster needs to process more image patches, resulting in longer generation times. Exploring ways to accelerate the generation process while maintaining image quality remains a valuable direction for future research in this field.
\section{Conclusion}
In this paper, we introduce ResMaster, a training-free, patch-based diffusion model for high-resolution image generation. ResMaster employs structural and fine-grained guidance strategies, ensuring the overall structural accuracy of high-resolution images while also achieving reasonable and rich local details. ResMaster can generate images of any scale and aspect ratio, and it offers faster inference speed compared to previous methods. Extensive experiments have demonstrated the superior capabilities of ResMaster.


{\small
\bibliographystyle{plain}
\bibliography{ref}
}


\newpage

\appendix

\section{Appendix / supplemental material}
\subsection{Implementation Details}
This subsection introduces some hyperparameter settings of our ResMaster. In the structural guidance module, we set the normalized stop frequency $D_0$ to 0.8 to ensure its control over the object structures during the generation process. Within the fine-grained guidance module, we utilize IP-Adapter~\cite{ye2023ip-adapter} to inject image prompts into SDXL~\cite{podell2023sdxl} as the condition. IP-Adapter~\cite{ye2023ip-adapter} is capable of generating images corresponding to the content of the image prompts. In this paper, we set the weight of the image prompt $\lambda$ to 0.8. Our framework is built on the patch-based diffusion model. We follow~\cite{du2024demofusion} to partition the entire noise into patches with a specified size [1024, 1024] and stride [64,64].
\subsection{Quantitative Results of Ablation Study}
The quantitative results of the ablation study are shown in \tablename~\ref{tab:quantitative_ablation}. All the results are performed in the resolution of $4096 \times 4096$.
\begin{table}[htpb]
\small
\centering
\caption{\textbf{Quantitative comparison results of the ablation study.} The best results are marked in \textbf{bold}. Base refers to the model that does not include the method proposed in this paper, namely Multi-Diffusion. Three components used in ResMaster: Structural Guidance (SG),
Tailored Image Prompts (TIP) and Enriched Textual Prompts (ETP).}
\label{tab:quantitative_ablation}
\begin{adjustbox}{width=0.9\linewidth,center}
\begin{tabular}{l|lllll}
\toprule[1pt]
Method & FID$_r$ $\downarrow$ & IS$_r$ $\uparrow$ & FID$_c$ $\downarrow$ & IS$_c$ $\uparrow$ & CLIP $\uparrow$ \\ \hline
Base (Multi-Diffusion) &  92.80   &  9.33   & 68.8  &  13.63   &  27.19    \\
Base+SG                &  79.43    & 14.59    &  64.24    & 15.79    & 28.39     \\
Base+SG+TIP            &  66.36    & 18.14    & 56.86     & 16.33    &  30.28    \\
Base+SG+TIP+ETP (ResMaster)   & \textbf{65.43}      & \textbf{18.44}     &  \textbf{55.09}   & \textbf{16.51}     & \textbf{30.95}         \\ \midrule[1pt]
\end{tabular}
\end{adjustbox}
\end{table}
\subsection{More Visualizations}
\begin{figure}[h]
    \centering
    \includegraphics[width=1.\linewidth]{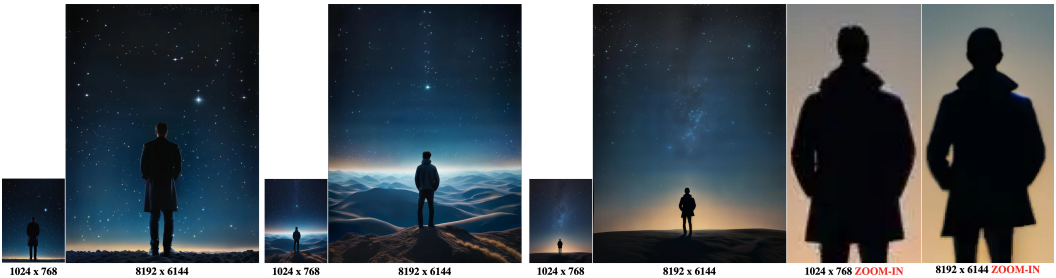}
    \vspace{-3mm}
    \caption{The comparisons of image quality between ResMaster and SDXL.}
    \label{fig:appendix0}
\end{figure}

\begin{figure}[h]
    \centering
    \includegraphics[width=1.\linewidth]{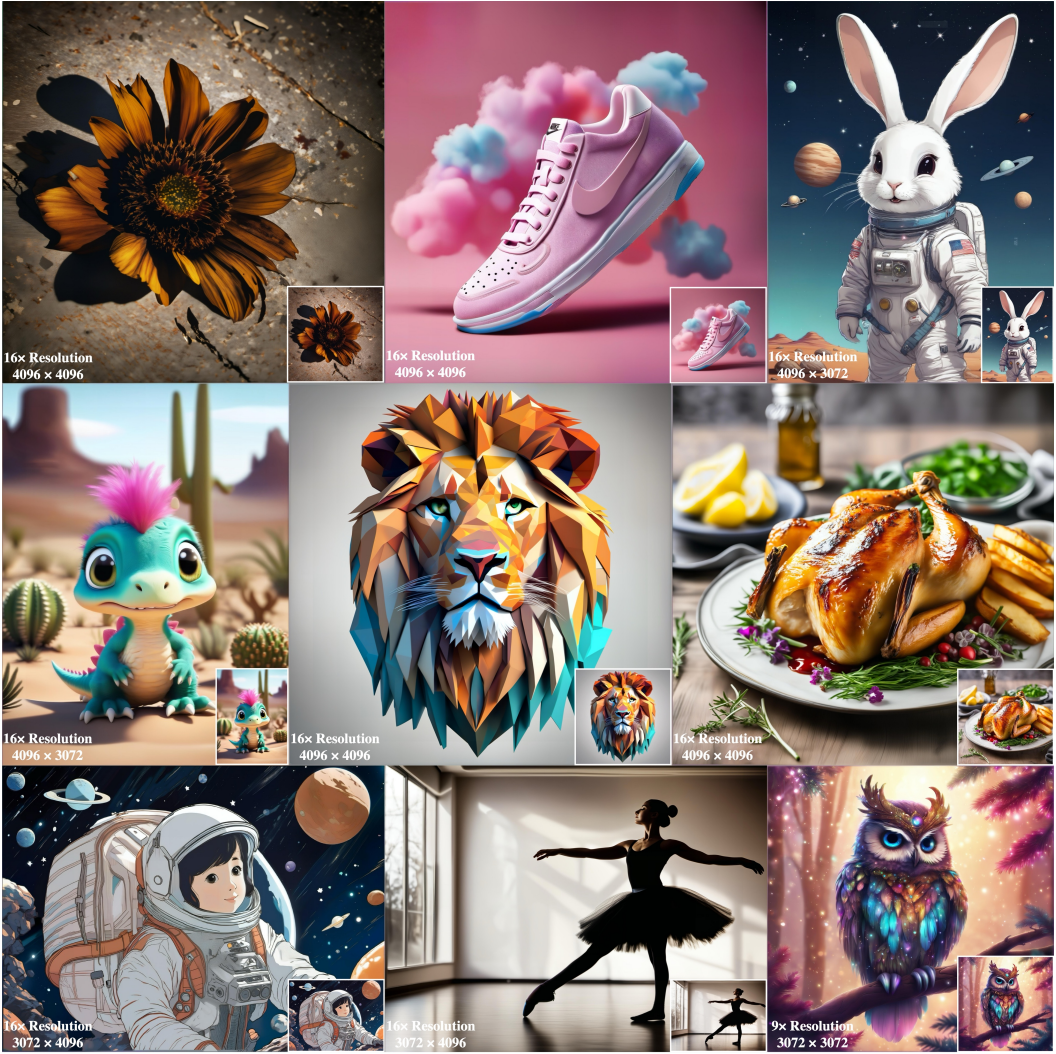}
    \vspace{-3mm}
    \caption{\textbf{More cases generated by ResMaster versus SDXL~\cite{podell2023sdxl}.} \textbf{Best viewed ZOOMED-IN.}}
    \label{fig:appendix_qual}
\end{figure}

\begin{figure}[h]
    \centering
    \includegraphics[width=1.\linewidth]{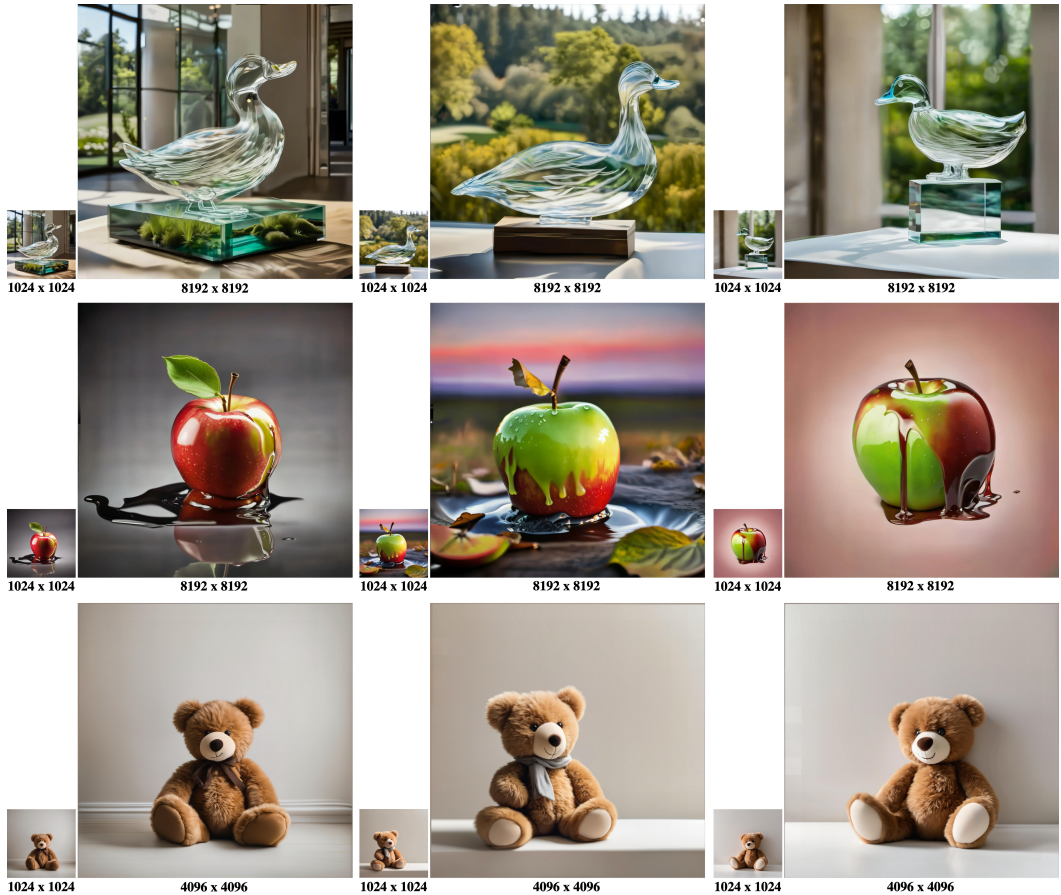}
    \vspace{-3mm}
    \caption{\textbf{More cases generated by ResMaster versus SDXL~\cite{podell2023sdxl}.} ResMaster can further upscale the generated results by 64 times or more without retraining the text-to-image diffusion model, maintaining high quality. \textbf{Best viewed ZOOMED-IN.}}
    \label{fig:appendix1}
\end{figure}

In \figurename~\ref{fig:appendix0}, \figurename~\ref{fig:appendix_qual} and \figurename~\ref{fig:appendix1}, we supplement more cases and comparisons to show the performance of ResMaster. In particular, we further show the results of $8192 \times 6144$ and $8192 \times 8192$ ($64\times$ upscale). These images clearly show that our $8K$ results still possess high detail retention and clarity, further demonstrating the capability of ResMaster.

\subsection{Prompts Used in This Paper}
\noindent\textbf{\figurename~\ref{fig:teaser} in the main text:}
\begin{itemize}
    \item \textit{A quiet library room, shelves stocked with books and a lone reader immersed in a novel by the window.}
\end{itemize}
\noindent\textbf{\figurename~\ref{fig:teaser2} in the main text:}
\begin{itemize}
    \item \textit{A Epic Rabbit made of Lego Technics with a few piece of Lego on the Table, Award winning, trending on ArtStation , (intricate details, masterpiece, best quality), (Use Dream Diffusion Secret Prompt), (Framed 3D), UHD}
    \item \textit{A loyal robotic dog with sleek, cyber-enhanced features standing guard beside its owner in a neon-lit cyberpunk city, bond depicted amidst a chaotic megacity at dusk, painted by Hajime Sorayama and Greg Rutkowski, sharp details and dynamic lighting.}
    \item \textit{A middle-aged woman of Asian descent, her dark hair streaked with silver, appears fractured and splintered, intricately embedded within a sea of broken porcelain. The porcelain glistens with splatter paint patterns in a harmonious blend of glossy and matte blues, greens, oranges, and reds, capturing her dance in a surreal juxtaposition of movement and stillness. Her skin tone, a light hue like the porcelain, adds an almost mystical quality to her form.}
    \item \textit{A bowl of fresh, vibrant cherries with water splashing around, high contrast, detailed, photorealistic, dark background.}
    \item \textit{a mystical tribal goddess adorned with feathers and gemstones and cables and synthesizer parts is surrounded by sacred geometry made from elven architecture, full body, gorgeous, perfect face, powerful, cinematic, beautifully lit, by artgerm, by karol bak, 3 d, trending on artstation, octane render, 8 k}
    \item \textit{A car made out of vegetables.}
    \item \textit{titanfall mech standing with its human pilot, dramatic lighting, illustration by greg rutkowski, yoji shinkawa, 4 k, digital art, concept art, trending on artstation}
    \item \textit{A vintage bicycle leaning against a rustic brick wall in an old European alley, flowers in its basket, painted in an impressionistic style by James Jean, capturing the charm of a bygone era.}
    \item \textit{4 k 8 k photorealistic torso headshot portrait of elegant asuka langley in red - white tight fit contact suit reflective by james jean, zac retz, pixiv}
    \item \textit{A band of musicians performing on stage at a lively concert. The stage is lit with colorful lights, and the crowd is cheering.}
    \item \textit{The mesmerizing northern lights dancing in the night sky over a frozen lake. The ice reflects the vibrant colors of the aurora borealis, adding to the surreal beauty of the scene.}
    \item \textit{a powerful energy woman, by alexander fedosav, hyper detailed digital matte painting, concept art, hyperrealism, 1 6 k resolution, cinema 4 d, 8 k resolution, trending on artstation, behance hd, a masterpiece, by stephan martiniere, particles, cel - shaded, power bright neon energy, by david a. hardy}
    \item \textit{Japanese Ukiyo-e, Kanagawa Surfing Sato.}
    \item \textit{cyborg sweating water, big drops of sweat, forehead only, by Hajime Sorayama, airbrush art, beautiful face, highly realistic, star flares, trending on artstation, beautiful lighting, sharp, details, hyper-detailed, HD, HDR, 4K, 8K.}
    \item \textit{A cozy winter scene, a snow-covered cabin with smoke rising from the chimney against a backdrop of pine trees, best quality, 4K.}
    \item \textit{A dense forest shrouded in mist, with towering pine trees and a narrow path winding through. Sunlight filters through the fog, creating a mystical, ethereal atmosphere.}
\end{itemize}

\noindent\textbf{\figurename~\ref{fig:ablation} in the main text:}
\begin{itemize}
    \item \textit{Pirate ship trapped in a cosmic maelstrom nebula, rendered in cosmic beach whirlpool engine, volumetric lighting, spectacular, ambient lights, light pollution, cinematic atmosphere, art nouveau style, illustration.}
\end{itemize}

\noindent\textbf{\figurename~\ref{fig:appendix0} in the main text:}
\begin{itemize}
    \item \textit{8k uhd A man looks up at the starry sky, lonely and ethereal, Minimalism, Chaotic composition Op Art.}
\end{itemize}

\noindent\textbf{\figurename~\ref{fig:appendix_qual} in the main text:}
\begin{itemize}
    \item \textit{mkitdecy, sad rusty decayed broken flower, deep shadows, high quality, high resolution, cinematic, dark}
    \item \textit{Nike sneaker concept art, (((made out of cotton candy clouds))) , luxury, futurist, stunning unreal engine render, product photography, 8k, hyper-realistic. Surrealism}
    \item \textit{cute rabbit in a spacesuit}
    \item \textit{an adorable and fluffy baby dinosaur with big color eyes, with soft feathers and wings, in the desert with cactus, with blur background, high quality, 8k}
    \item \textit{a lion, colorful, low-poly, cyan and orange eyes, poly-hd, 3d, low-poly game art, polygon mesh, jagged, blocky, wireframe edges, centered composition}
    \item \textit{rotisserie chicken dish, culinary delight 8k insta}
    \item \textit{a girl astronaut exploring the cosmos, floating among planets and stars, high quality detail, , anime screencap, studio ghibli style, illustration, high contrast, masterpiece, best quality}
    \item \textit{beautiful silhouette shot of a ballerina dancer}
    \item \textit{Happy dreamy owl monster sitting on a tree branch, colorful glittering particles, forest background, detailed feathers.}
\end{itemize}

\noindent\textbf{\figurename~\ref{fig:appendix1} in the main text:}
\begin{itemize}
    \item \textit{A transparent sculpture of a duck made out of glass. The sculpture is in front of a painting of a landscape.}
    \item \textit{a melting apple.}
    \item \textit{A cute teddy bear in front of a plain white wall, warm and brown fur, soft and fluffy.}
\end{itemize}


\end{document}